\documentclass{article}
\usepackage{spconf,amsmath,graphicx}
\usepackage{amsmath,amsfonts}
\usepackage{algorithmic}
\usepackage{algorithm}
\usepackage{array}
\usepackage[caption=false,font=normalsize,labelfont=sf,textfont=sf]{subfig}
\usepackage{textcomp}
\usepackage{stfloats}
\usepackage{url}
\usepackage{verbatim}
\usepackage{graphicx}
\usepackage{cite}
\usepackage{picture}
\usepackage{verbatim}
\usepackage{pgfplots}
\usepackage{booktabs}
\usepackage{multicol}
\usepackage{multirow}
\usepackage{xcolor}

\newcommand\red{\textcolor{red}}
\newcommand\blue{\textcolor{blue}}

\title{Encoder-minimal and Decoder-minimal Framework for Remote Sensing Image Dehazing}

\begin{document}
%\ninept
%

\name{Yuanbo Wen$^{1}$ \qquad Tao Gao$^{1*}$ \qquad Ziqi Li$^{1}$ \qquad Jing Zhang$^{2}$ \qquad Ting Chen$^{1*}$}
\address{$^{1}$ School of Information Engineering, Chang'an University, Xi'an, China; $^{2}$ College of Engineering
\\ and Computer Science, Australian National University, Canberra, ACT, Australia}

\maketitle
\begin{abstract}
Haze obscures remote sensing images, hindering valuable information extraction.
To this end, we propose RSHazeNet, an encoder-minimal and decoder-minimal framework for efficient remote sensing image dehazing.
Specifically, regarding the process of merging features within the same level, we develop an innovative module called intra-level transposed fusion module (ITFM). This module employs adaptive transposed self-attention to capture comprehensive context-aware information, facilitating the robust context-aware feature fusion.
Meanwhile, we present a cross-level multi-view interaction module (CMIM) to enable effective interactions between features from various levels, mitigating the loss of information due to the repeated sampling operations.
In addition, we propose a multi-view progressive extraction block (MPEB) that partitions the features into four distinct components and employs convolution with varying kernel sizes, groups, and dilation factors to facilitate view-progressive feature learning.
Extensive experiments demonstrate the superiority of our proposed RSHazeNet.
We release the source code and all pre-trained models at \url{https://github.com/chdwyb/RSHazeNet}.
\end{abstract}
\section{Introduction}
\label{sec:intro}

\footnotetext[1]{Corresponding author: Tao Gao, Ting Chen}
\footnotetext[2]{This research was partially supported by the National Key R \& D Program of China under Grants 2019YFE0108300, the National Natural Science Foundation of China under Grants 52172379, 62001058 and U1864204, the Fundamental Research Funds for the Central University under Grants 300102242901.}

Haze emerges as a prevalent meteorological phenomenon significantly compromising remote sensing image quality. It leads to a reduction in image contrast and color fidelity, subsequently impeding various observation applications such as image classification \cite{gao2022heavy}, modality translation \cite{liu2021modality}, and image super-resolution \cite{yue2022super}.
Consequently, remote sensing image dehazing holds practical significance in enhancing the quality of captured images affected by haze conditions.

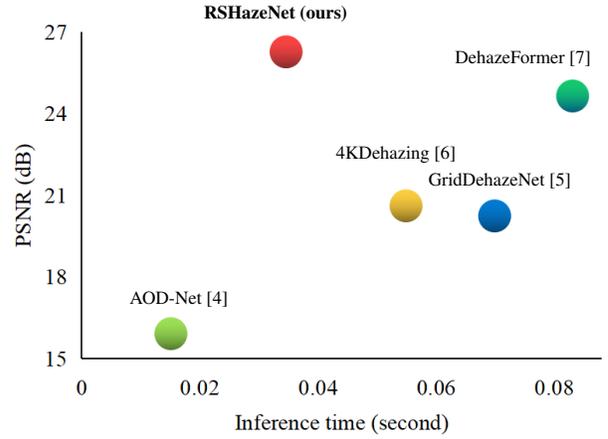
\begin{figure}[!t]
    \centering
    \hspace{-50mm}
    \begin{picture}(100,170)
    \put(0,0){\includegraphics[width=\linewidth]{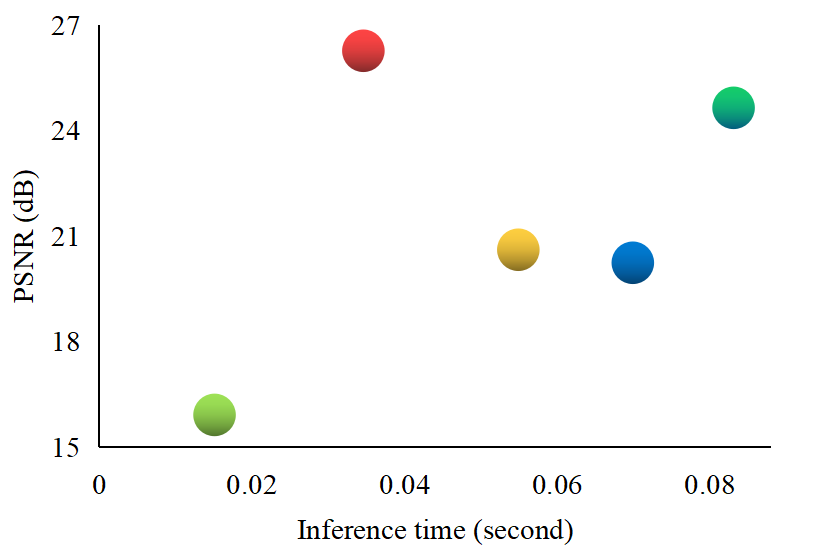}}
    \put(47,52){\footnotesize{\scriptsize AOD-Net \cite{li2017aod}}}
    \put(160,97){\footnotesize{\scriptsize GridDehazeNet \cite{valanarasu2022transweather}}}
    \put(125,107){\footnotesize{\scriptsize 4KDehazing \cite{xiao2022single}}}
    \put(170,143){\footnotesize{\scriptsize DehazeFormer \cite{song2023vision}}}
    \put(75,160){\footnotesize{\textbf{\scriptsize RSHazeNet (ours)}}}
    \end{picture}
    \vspace{-3mm}
    \caption{Comparative analysis of efficiency in remote sensing image dehazing approaches. Our novel approach, RSHazeNet, exhibits superior performance in terms of both metrical scores and computational efficiency.}
    \label{fig:efficiency}
    \vspace{-4mm}
\end{figure}

Remote sensing image is always in large resolution while the intensity and distribution of haze are more various.
Therefore, most existing dehazing methods \cite{song2023vision, hoang2023transer, li2022m2scn, li2017aod, shao2020domain, tu2022maxim} may not be well-suited for remote sensing image dehazing. In response, DADN \cite{gu2019single} introduces a prior-based dense attentive network that leverages dense blocks and attention blocks. FCTF-Net \cite{li2020coarse} adopts a two-stage framework, while DCRD-Net \cite{huang2021single} employs a dual-step cascaded residual dense network to directly reconstruct the background.
Additionally, H2RL-Net \cite{chen2021hybrid} proposes a multi-scale architecture to mitigate haze degradation. M2SCN \cite{li2022m2scn} incorporates a multi-model joint estimation module and a self-correcting module to construct an end-to-end network. More recently, DehazeFormer \cite{song2023vision} introduces a modified normalization layer and a spatial information aggregation scheme.
However, these methods face challenges in striking the optimal balance between image quality and inference time.

\begin{figure}[!t]
    \centering
    \includegraphics[width=\linewidth]{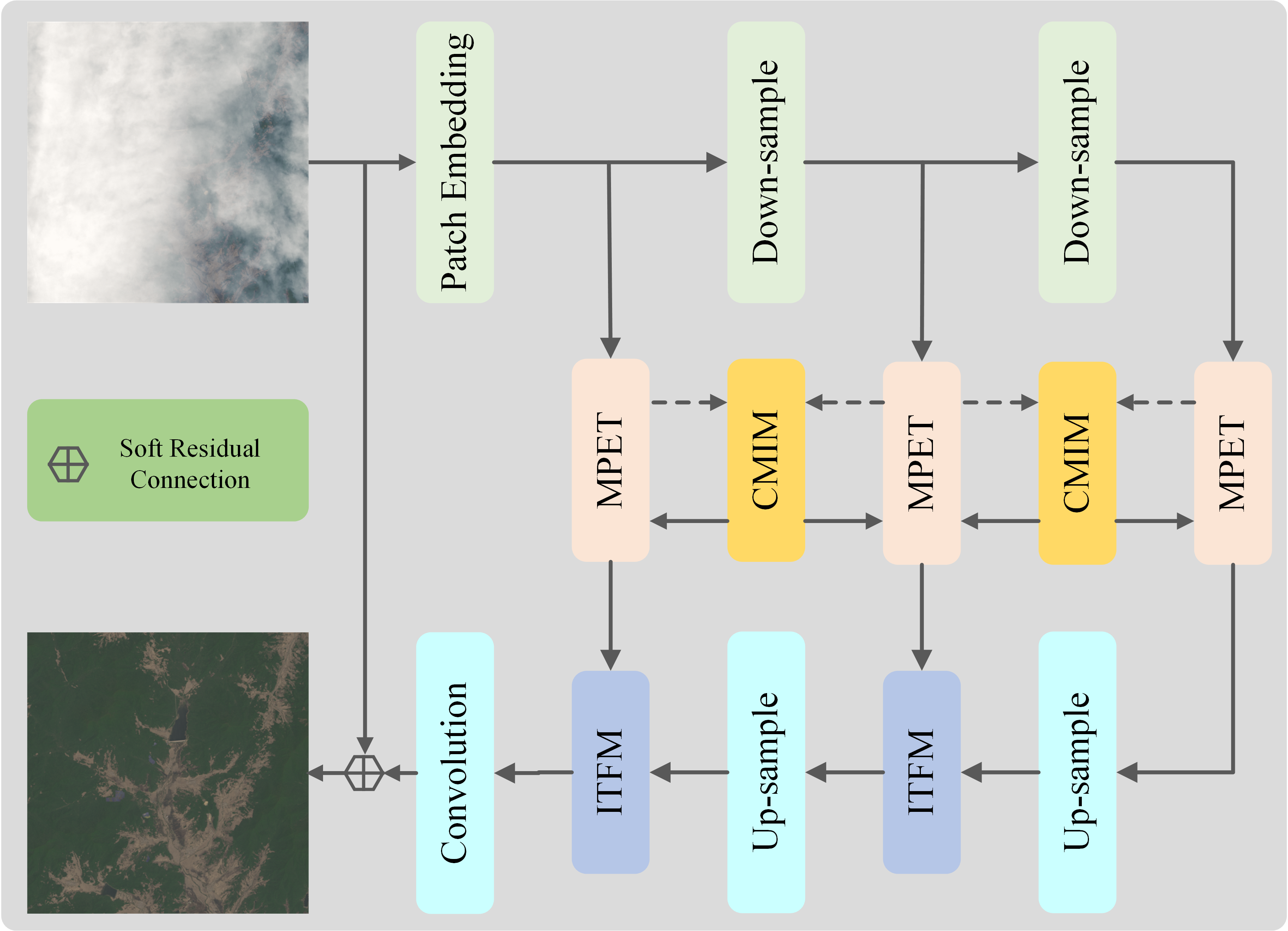}
    \caption{Architectural overview of our proposed encoder-minimal and decoder-minimal RSHazeNet for remote sensing image dehazing.}
    \label{fig:network}
\end{figure}

In this work, we redirect our attention towards skip connections and propose an encoder-minimal and decoder-minimal approach.
As depicted in Figure \ref{fig:network}, the encoding and decoding stages solely consist of down-sampling and up-sampling operations.
Within u-shaped structures, we introduce a intra-level transposed fusion module (ITFM), which incorporates adaptive pooling to capture global spatial information and leverages efficient transposed self-attention to compute attention maps, amalgamating information from distinct feature planes and extracting context-wise attention.
By adopting ITFM, we significantly mitigate the computational complexity associated with self-attention and effectively harness global enhanced information to fuse features.

Existing methods exhibit a lack of capability to establish interactions across different levels, resulting in the hindered flow of information between multi-scale features.
To this end, we present a novel module called cross-level multi-scale interaction module (CMIM), which facilitates feature interaction between two features of varying resolutions. In CMIM, we employ transposed self-attention to compute attention maps \cite{gao2023towards}. Specifically, the query representation Q and key representation K are derived from the input features at level $l$ and level $l-1$, respectively. We generate two value representations, which are both enhanced by the same interacting attention map, thus enabling features to flow at different stages and avoid the information loss due to the repeated sampling operations.

Moreover, we introduce a multi-view progressive extraction block (MPEB). Unlike employing $3\times 3$ convolutions \cite{chen2023run}, MPEB exclusively utilizes channel-wise interactions accomplished through $1\times 1$ convolutions.
To capture the multi-view features from the remaining three quarters of the input, we incorporate grouped convolutions with distinct receptive fields and dilation factors.
By adopting MPEB, we not only achieve channel-wise information interaction, but also extract multi-view spatial-wise features. This allows the network to capture different viewpoints of the input features, providing a more robust and discriminative representation.

The overall architecture of our proposed RSHazeNet is depicted in Figure \ref{fig:network}, which achieves the best trade-off between image quality and inference time. We summarize our contributions as follows:
1) We devise a novel architectural design strategy that shifts the emphasis onto skip connections instead of encoders and decoders.
2) An intra-level transposed fusion module that incorporates global information to facilitate the comprehensive and context-aware fusion process.
3) A cross-level multi-scale interaction module that enables effective communication and collaboration between different levels, ensuring that valuable information is preserved and shared.
4) We present a multi-view progressive extraction block that accomplishes both channel-wise interaction and spatial-wise multi-view feature extraction.

\section{Proposed Method}

\subsection{Intra-level Transposed Fusion Module}

\begin{figure}[!t]
    \centering
    \includegraphics[width=\linewidth]{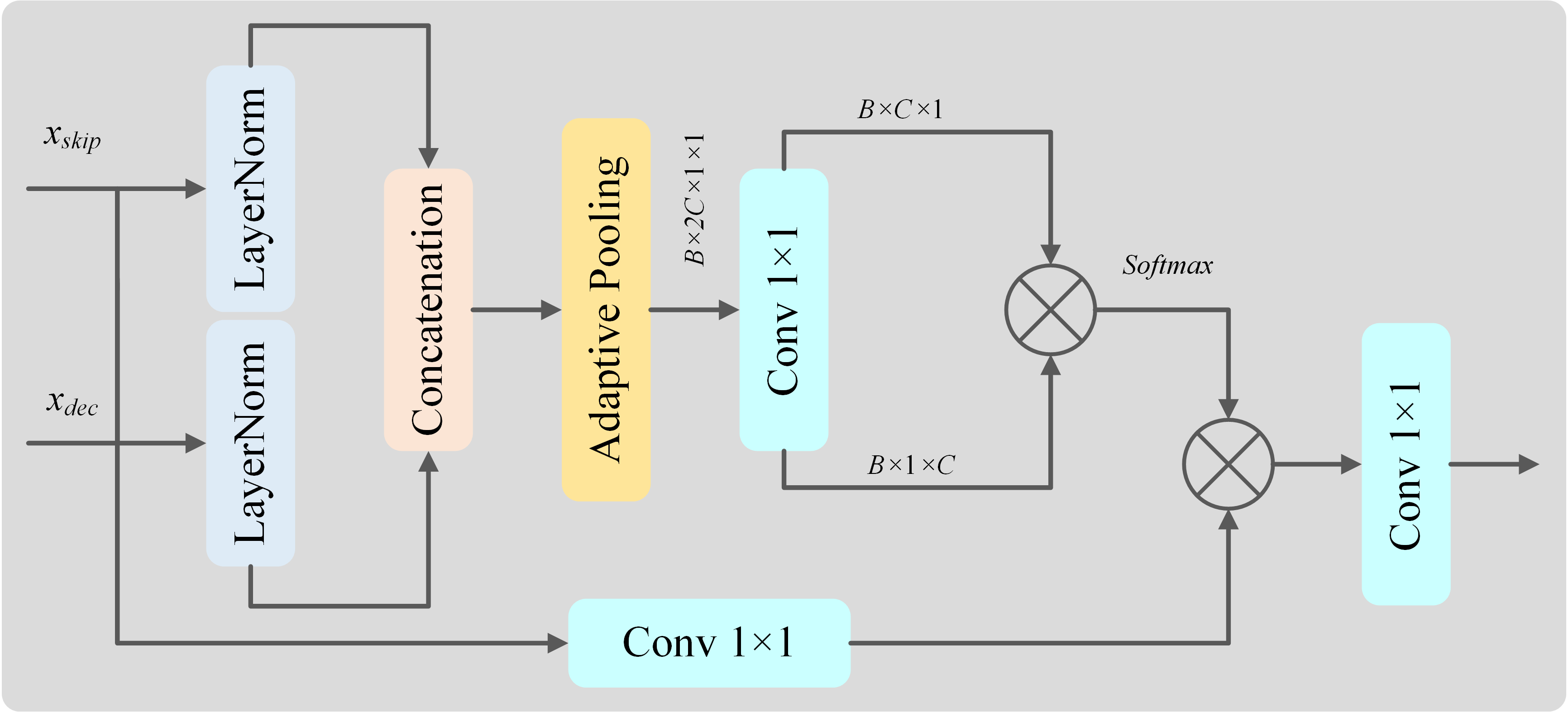}
    \caption{Illustration of our proposed intra-level transposed fusion module (ITFM). It utilizes the adaptive spatial features to calculate the channel-wise attention map and further conduct context-based fusion.}
    \label{fig:itfm}
    \vspace{-2mm}
\end{figure}

To overcome the limitations in effectively aggregating information from different feature planes, we propose an intra-level transposed fusion module (ITFM) as illustrated in Figure \ref{fig:itfm}. It leverages features from two different sources to represent attention.
In ITFM, to capture channel-wise attention, we employ adaptive pooling and $1\times 1$ convolution to generate the query representation Q and key representation K.
This design allows ITFM to effectively combine information from different sources and capture contextual dependencies.
Q and K are derived by
\begin{equation}
    Q, K = conv_{1\times 1}( pool(concat(LN(x_{skip}), LN(x_{dec})))),
\end{equation}
where the inputs from skip connection and up-sampling operator are $x_{skip}$ and $x_{dec}$, $LN$ denotes layer normalization, $concat$ signifies channel-wise concatenation, $pool$ represents adaptive pooling operation, while $conv_{1\times 1}$ denotes the $1\times 1$ convolution.
The attention map is formulated as
\begin{equation}
\label{eq:attention}
    A = softmax((Q^{{\rm T}}\otimes K ) \cdot \alpha),
\end{equation}
where $\otimes$ denotes the matrix multiplication, T indicates the matrix transposing operation, $\alpha$ is a trainable scaling parameter. Finally, we employ the two input features to represent value V. The fuses features can be expressed as
\begin{equation}
    x = conv_{1\times 1}(A\otimes (conv_{1\times 1}(concat(x_{skip}, x_{dec})))^{{\rm T}})
\end{equation}
Our representations Q and K are tailored to encompass singular global spatial features, leading to a substantial reduction in computational overhead.

\subsection{Cross-level Multi-scale Interaction Module}

\begin{figure}[h]
    \centering
    \includegraphics[width=\linewidth]{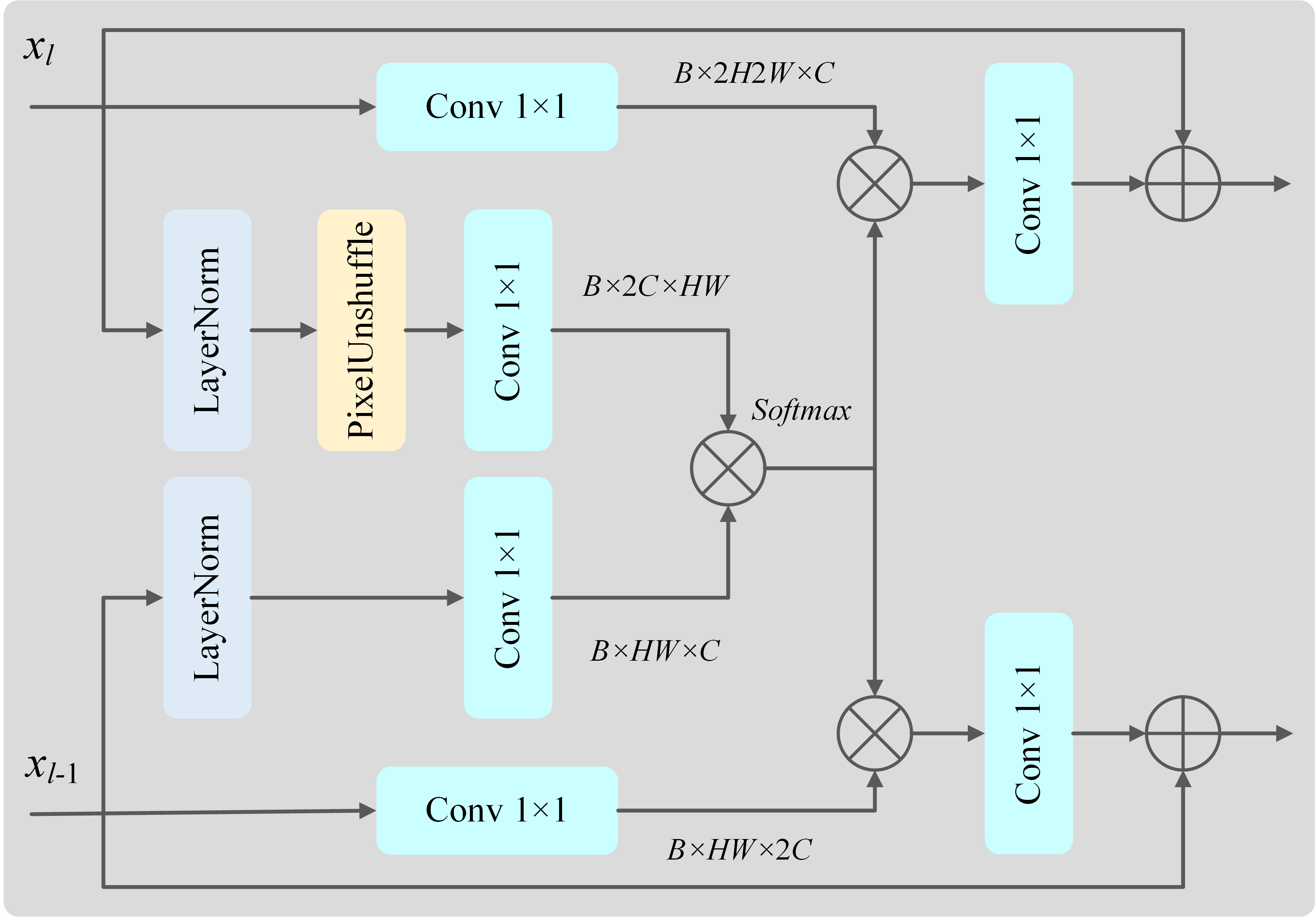}
    \caption{Illustration of our proposed cross-level multi-scale interaction module (CMIM) that facilitates features interaction from different levels.}
    \label{fig:cmim}
    \vspace{-2mm}
\end{figure}

To address the limitation that existing architectures \cite{song2023vision, wang2022dynamic, dong2022transra, bie2022single} lack the ability to establish effective interactions across different levels,
we propose a novel cross-level multi-scale interaction module (CMIM) to facilitate global interaction.
As depicted in Figure \ref{fig:cmim}, we unify the features and bring them to the same resolution. Subsequently, we obtain the transposed self-attention by performing matrix transposition. The value representations obtained from the skip connection and up-sampling operation are preserved in sizes identical to their corresponding input features.
In CMIM, Q and K are derived as
\begin{equation}
\begin{aligned}
    & Q = conv_{1\times 1}(PU(LN(x_{l}))), \\
    & K = conv_{1\times 1}(LN(x_{l-1})),
\end{aligned}
\end{equation}
where $PU$ denotes the pixel-unshuffle operation, $x_{l}$ and $x_{l-1}$ are the features from level $l$ and level $l-1$.
Moreover, the enhanced representations of $x_l$ and $x_{l-1}$ can be formulated as
\begin{equation}
    \begin{aligned}
        & x_{l}^{'} = x_l + conv_{1\times 1}(A\otimes conv_{1\times 1}(x_l)), \\
        & x_{l-1}^{'} = x_{l-1} + conv_{1\times 1}(A^{{\rm T}}\otimes conv_{1\times 1}(x_{l-1})),
    \end{aligned}
\end{equation}
where $x_{l}^{'}$ and $x_{l-1}^{'}$ denote the enhanced features.
Our CMIM offers two key advantages. Firstly, it addresses the issue of information loss that commonly occurs due to the repeated up-sampling and down-sampling. Secondly, the multi-scale features obtained from one level play a crucial role in enriching the features of the subsequent level.

\subsection{Multi-view Progressive Extraction Block}

\begin{figure}[h]
    \centering
    \includegraphics[width=\linewidth]{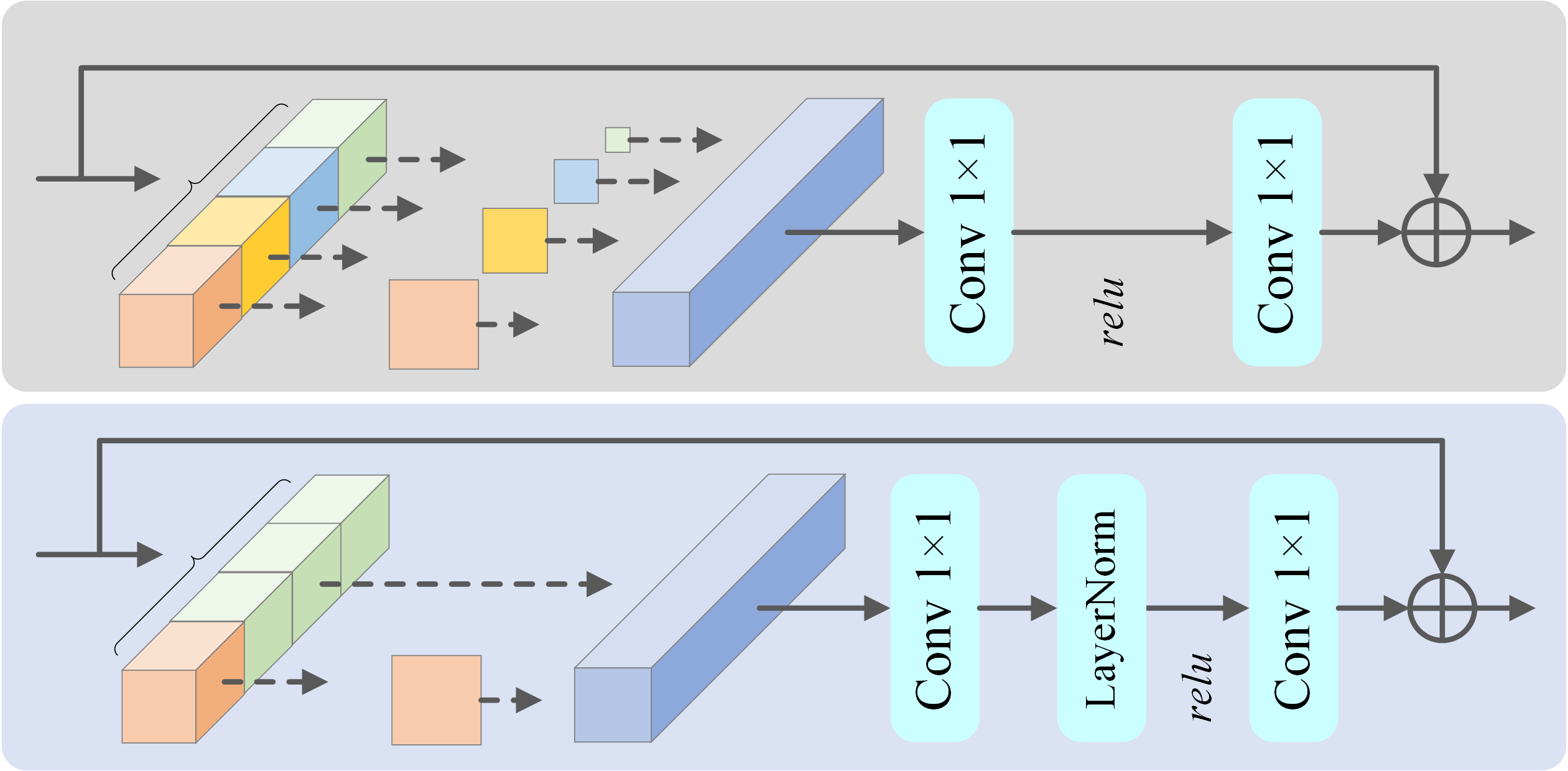}
    \caption{Illustration of the basic learning block. \textit{Top}: our proposed multi-view progressive extraction module. \textit{Bottom}: the FasterNetBlock (FNB) with one fourth partial convolution.}
    \label{fig:mpeb}
    \vspace{-2mm}
\end{figure}

Considering the high-resolution characteristics of remote sensing hazy images, we propose a novel multi-view progressive extraction block (MPEB).
As illustrated in Figure \ref{fig:mpeb}, three-quarters of the features are divided into three parts, each corresponding to a grouped dilated convolution with different receptive fields.
Although the sizes of the convolution kernels are not the same, they share the same dilation factors.
This choice of constant dilation factors avoids the grid problem.
The last one-quarter features are processed by only a 1$\times$1 convolution.
The formulation of our MPEB is
\begin{equation}
\begin{aligned}
    & x_{i}^{'} = conv_{2i-1\times 2i-1, d, g}(x_i), i=1, 2, 3, 4. \\
    & x = mlp(concat(x_{1}^{'}, x_{2}^{'}, x_{3}^{'}, x_{4}^{'})),
\end{aligned}
\end{equation}
where $conv_{2i-1\times 2i-1, d, g}$ denotes the convolution with kernel size of $2i-1\times 2i-1$, dilation factor of $d$ and group size of $g$. $d$ and $g$ conform to the follow
\begin{equation}
    d=\left\{\begin{aligned}
        & 1,\ \  i=1 \\
        & 3,\ \  otherwise
    \end{aligned}
    \right.
    ,
    g=\left\{\begin{aligned}
        & 1,\ \ \ \ i=1 \\
        & \frac{1}{4}c,\ \  otherwise
    \end{aligned}
    \right.
    ,
\end{equation}
where $c$ denotes the channel number of input features.

\begin{table*}[h]
    \centering
    \caption{Remote sensing image dehazing results on StateHaze1K and RS-Haze datasets \cite{huang2020single, song2023vision}. Our proposed RSHazeNet obtains the best performance on all these tesing remote sensing hazy datasets.}
    \label{tab:statehaze1k}
    \renewcommand\arraystretch{1}
    \resizebox{\linewidth}{!}{
        \begin{tabular}{lccccccccccccccc}
            \toprule
            \multirow{2}{*}{Method} & \multicolumn{3}{c}{StateHaze1K-thin} & \multicolumn{3}{c}{StateHaze1K-moderate} & \multicolumn{3}{c}{StateHaze1K-thick} & \multicolumn{3}{c}{StateHaze1K-average} & \multicolumn{3}{c}{RS-Haze} \\
            ~ & PSNR & SSIM & MSE & PSNR & SSIM & MSE & PSNR & SSIM & MSE & PSNR & SSIM & MSE & PSNR & SSIM & MSE \\
            \midrule
            DCP \cite{he2010single} & 13.15 & 0.7246 & 0.0473 & 9.780 & 0.5735 & 0.1072 & 10.25 & 0.5850 & 0.0942 & 11.06 & 0.6477 & 0.0832 & 17.86 & 0.7340 & 0.0201 \\
            AOD-Net \cite{li2017aod} & 19.54 & 0.8543 & 0.0113 & 20.10 & 0.8854 & 0.0101 & 15.92 & 0.7313 & 0.0265 & 18.52 & 0.8234 & 0.0159 & 27.09 & 0.8476 & 0.0032 \\
            GridDehazeNet \cite{liu2019griddehazenet} & 24.34 & 0.9155 & 0.0037 & 23.05 & 0.9274 & 0.0051 & 20.24 & 0.8312 & 0.0096 & 22.54 & 0.8916 & 0.0061 &35.23 & 0.9446 & 0.0004 \\
            FCFT-Net \cite{li2020coarse} & 23.59 & 0.9127 & 0.0044 & 22.88 & 0.9272 & 0.0055 & 20.03 & 0.8156 & 0.0101 & 22.17 & 0.8852 & 0.0067 & 33.28 & 0.9417 & 0.0007 \\
            H2RL-Net \cite{chen2021hybrid} & 20.91 & 0.8797 & 0.0083 & 22.34 & 0.9061 & 0.0065 & 17.41 & 0.7684 & 0.0228 & 20.22 & 0.8514 & 0.0125 & 31.18 & 0.9212 & 0.0019 \\
            4KDehazing \cite{xiao2022single} & 22.35 & \blue{0.9550} & 0.0059 & 23.73 & 0.9565 & 0.0047 & 20.61 & \blue{0.9264} & 0.0088 & 22.23 & 0.9460 & 0.0065 & 33.80 & 0.9442 & 0.0007  \\
            M2SCN \cite{li2022m2scn} & 25.21 & 0.9175 & 0.0031 & 26.11 & 0.9416 & 0.0038 & 21.33 & 0.8289 & 0.0095 & 24.22 & 0.8960 & 0.0055 & 37.75 & 0.9497 & 0.0003 \\
            DehazeFormer \cite{song2023vision} & \blue{26.28} & 0.9477 & \blue{0.0024} & \blue{29.78} & \blue{0.9677} & \blue{0.0012} & \blue{24.64} & 0.9241 & \blue{0.0034} & \blue{26.90} & \blue{0.9465} & \blue{0.0023} & \blue{39.57} & \blue{0.9701} & \blue{0.0002}  \\
            \textbf{RSHazeNet} & \red{28.38} & \red{0.9666} & \red{0.0015} & \red{30.89} & \red{0.9717} & \red{0.0009} & \red{26.26} & \red{0.9325} & \red{0.0024} & \red{28.51} & \red{0.9569} & \red{0.0016} & \red{40.61} & \red{0.9754} & \red{0.0001} \\
            \bottomrule
        \end{tabular}
    }
\end{table*}

\section{Experiments}

\subsection{Implementation Specifications}
RSHazeNet is trained for a total of 1, 000 epochs using a single NVIDIA GeForce RTX 3090 GPU with patch size and batch size of $512\times 512$ and 14, respectively.
We utilize the Adam optimizer with an initial learning rate of 0.0002, which is gradually decreased to $1\times 10^{-8}$ using cosine annealing decay. 
During training, we apply data augmentation techniques such as random rotation and horizontal flipping.

\subsubsection{Datasets}

StateHaze1K dataset \cite{huang2020single} comprises three subsets: StateHaze1K-thin, StateHaze1K-moderate and StateHaze1K-thick. Each subset consists of 400 image pairs, with 320 designated for training, 35 for validation, and 45 for testing purposes.
RS-Haze dataset \cite{song2023vision} is a large-scale dataset containing 51300 paired images for training and 2700 paired images for testing.

\subsection{Comparisons to Existing Methods}

\begin{figure}[h]
\centering
    \includegraphics[width=1.3cm]{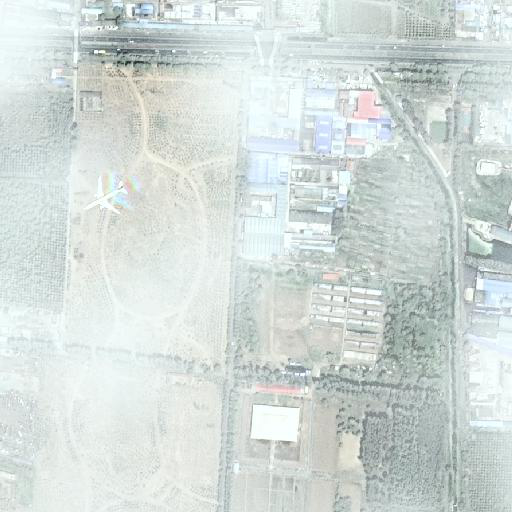}
    \includegraphics[width=1.3cm]{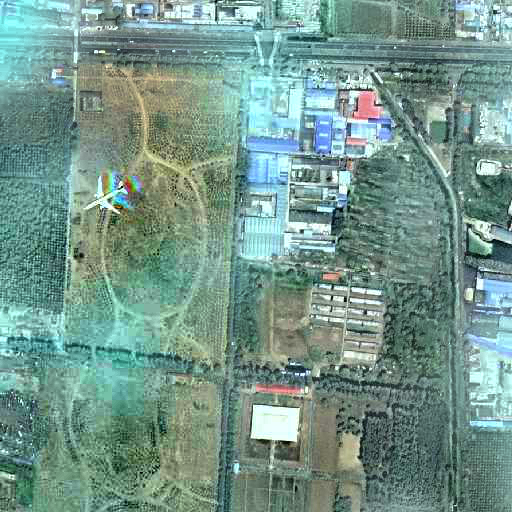}
    \includegraphics[width=1.3cm]{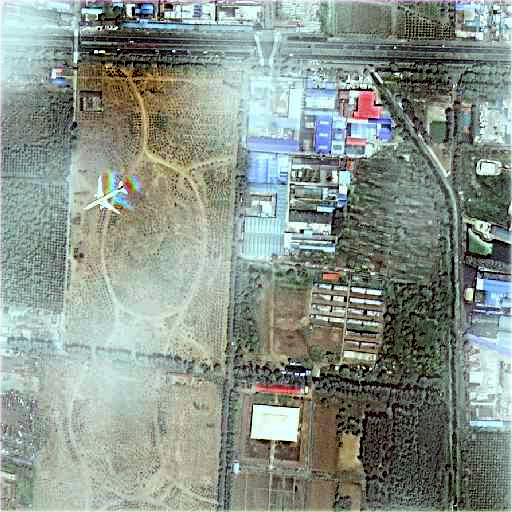}
    \includegraphics[width=1.3cm]{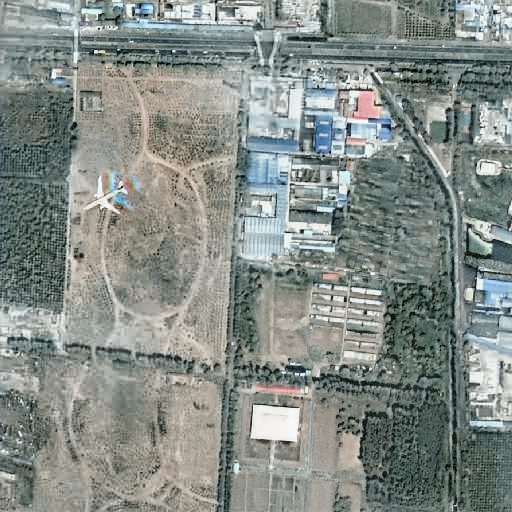}
    \includegraphics[width=1.3cm]{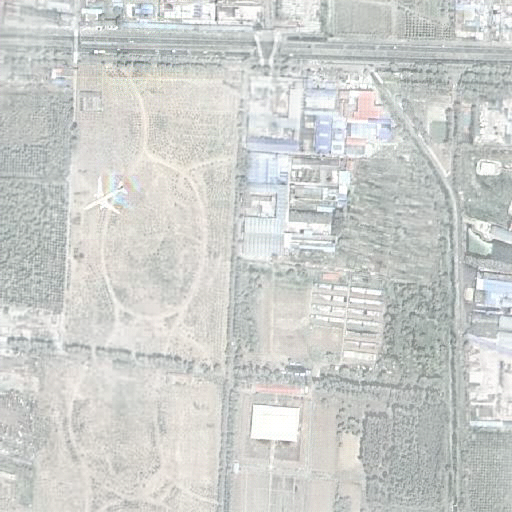}
    \includegraphics[width=1.3cm]{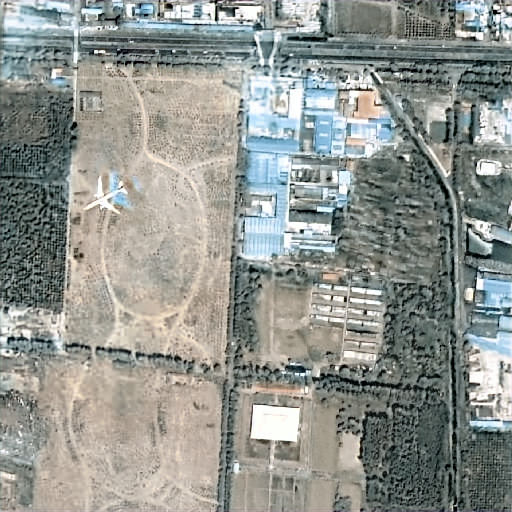}

    \vspace{1mm}
    \includegraphics[width=1.3cm]{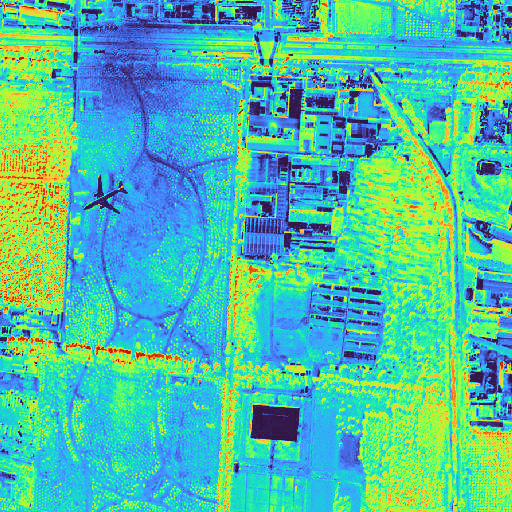}
    \includegraphics[width=1.3cm]{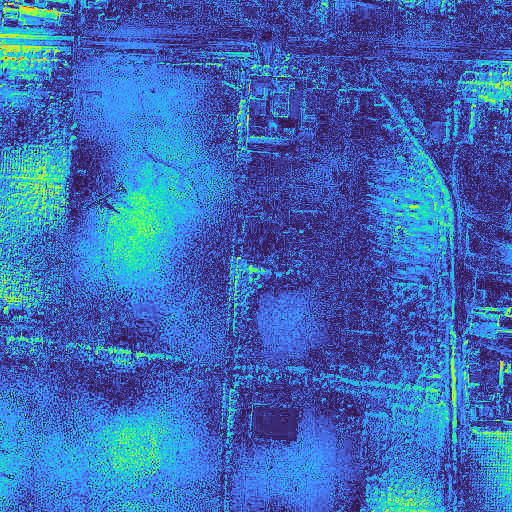}
    \includegraphics[width=1.3cm]{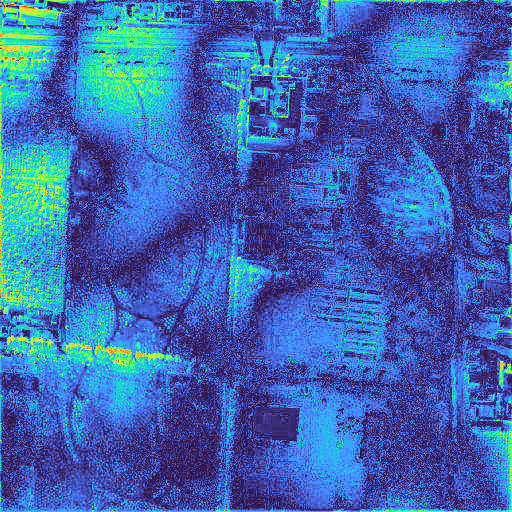}
    \includegraphics[width=1.3cm]{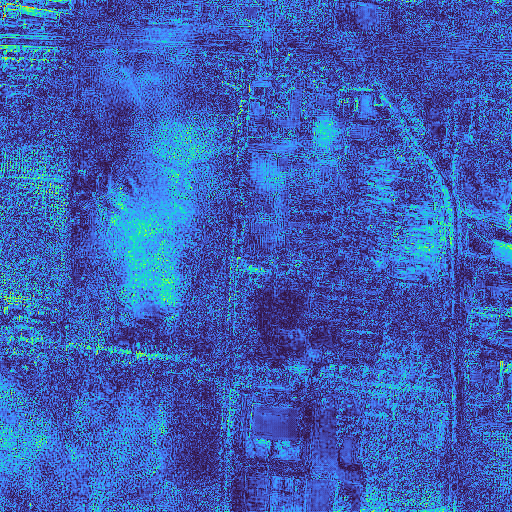}
    \includegraphics[width=1.3cm]{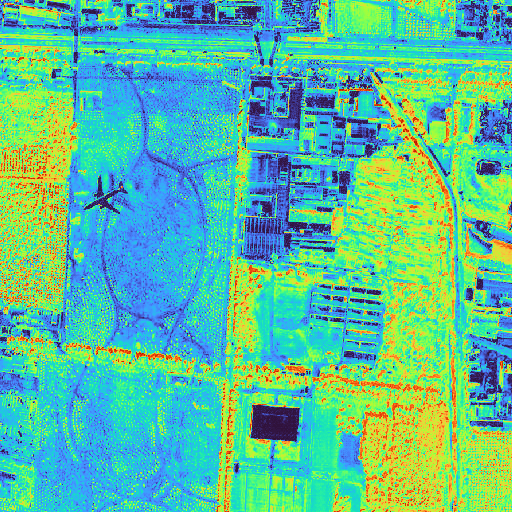}
    \includegraphics[width=1.3cm]{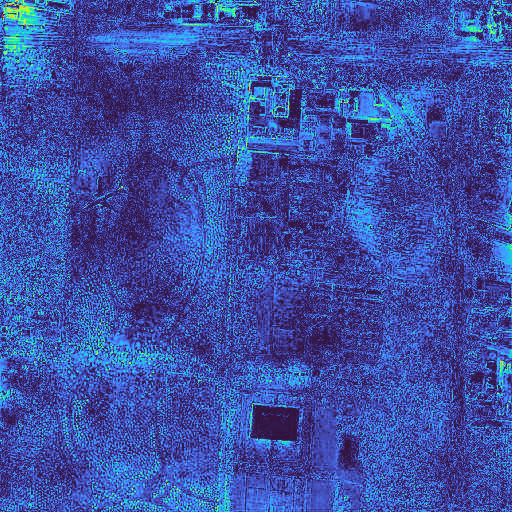}

    \vspace{1mm}
    \includegraphics[width=1.3cm]{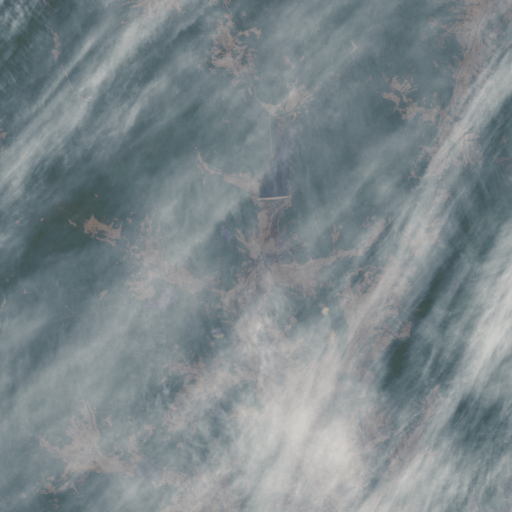}
    \includegraphics[width=1.3cm]{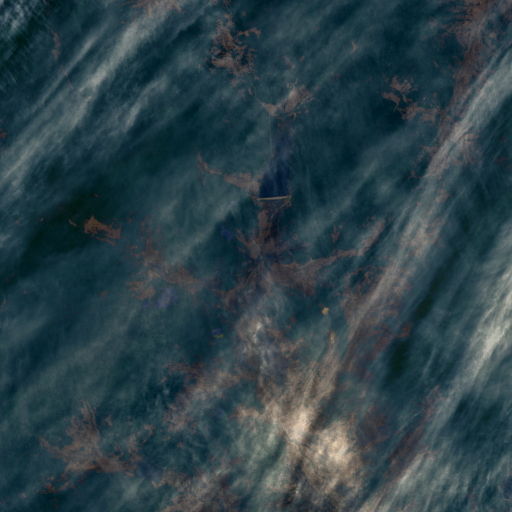}
    \includegraphics[width=1.3cm]{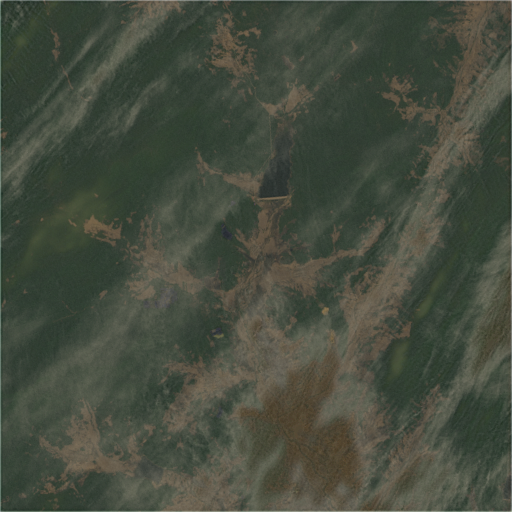}
    \includegraphics[width=1.3cm]{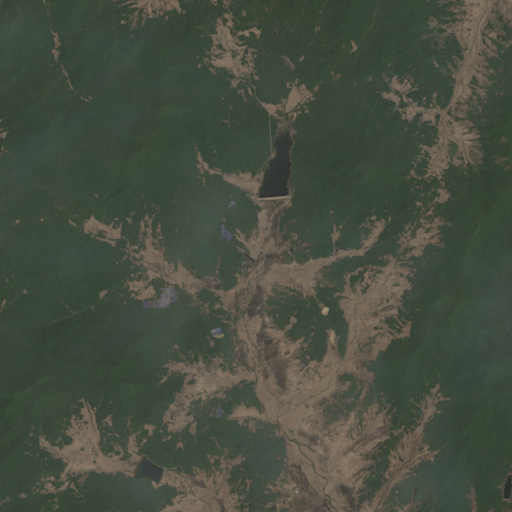}
    \includegraphics[width=1.3cm]{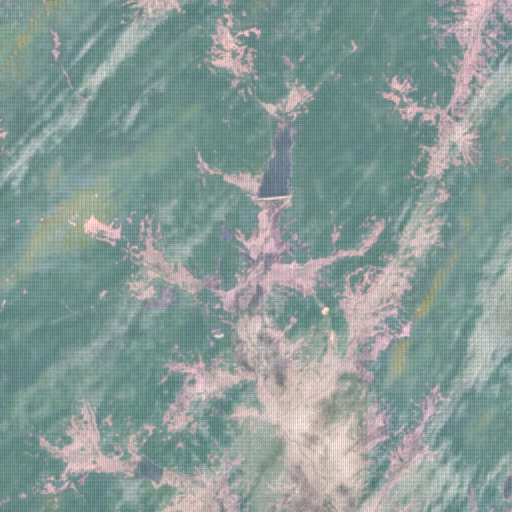}
    \includegraphics[width=1.3cm]{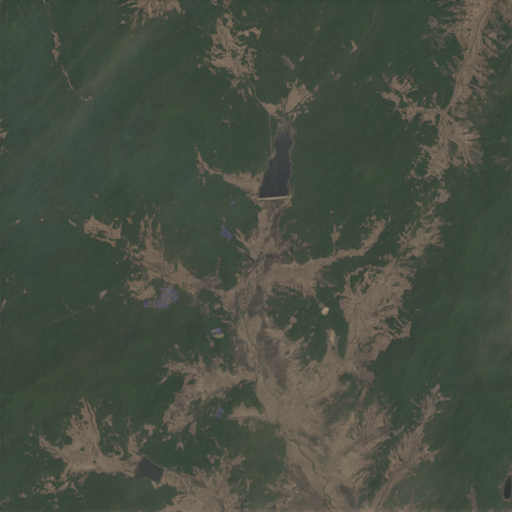}

    \vspace{1mm}
    \includegraphics[width=1.3cm]{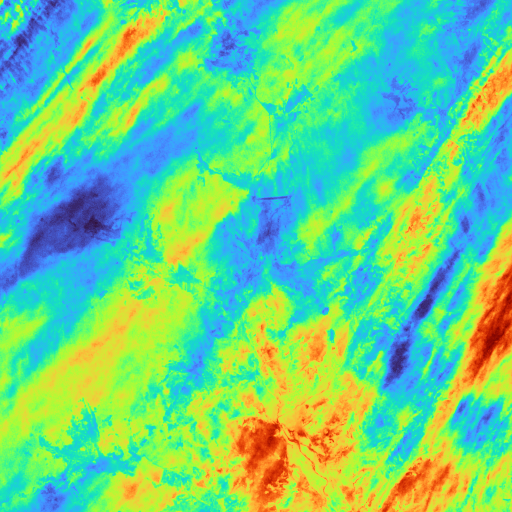}
    \includegraphics[width=1.3cm]{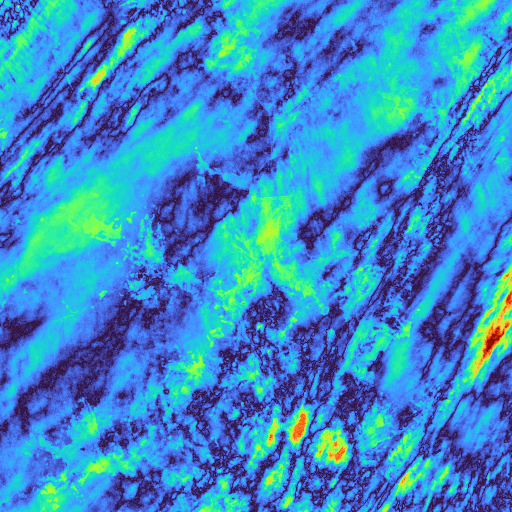}
    \includegraphics[width=1.3cm]{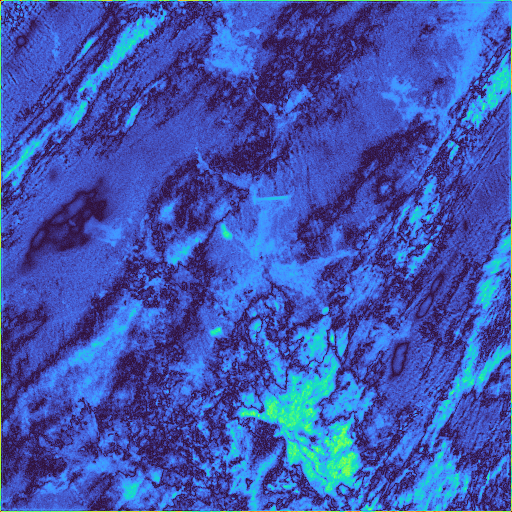}
    \includegraphics[width=1.3cm]{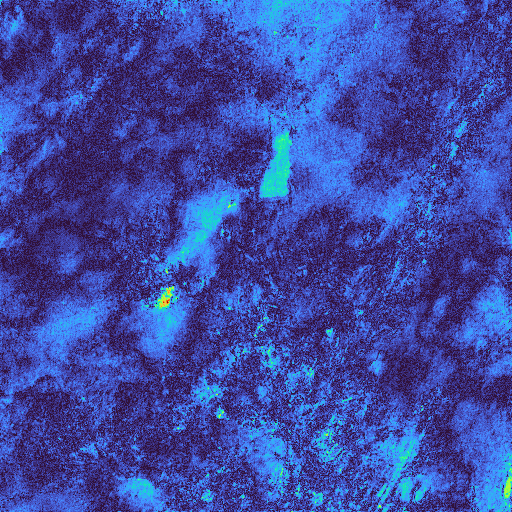}
    \includegraphics[width=1.3cm]{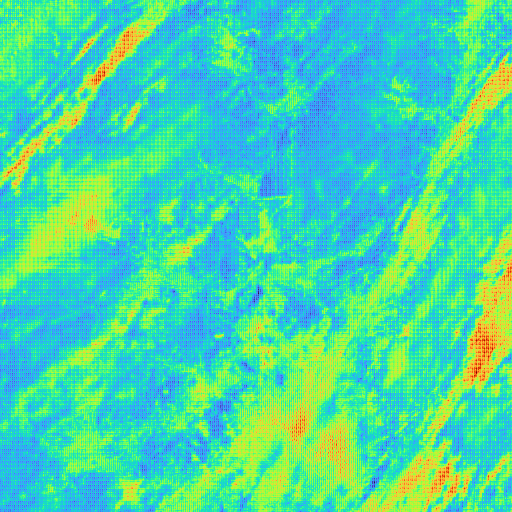}
    \includegraphics[width=1.3cm]{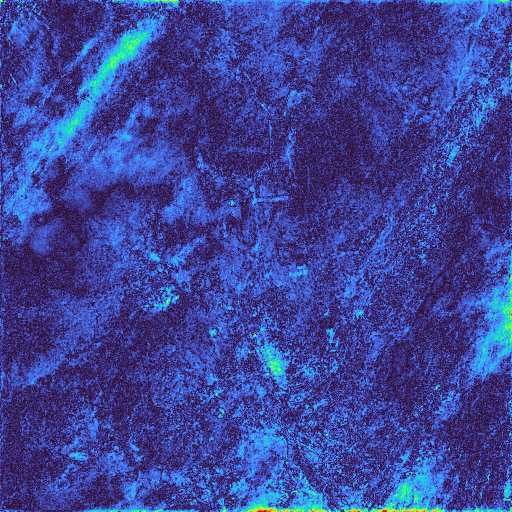}
    
    \caption{Visual comparisons of input iamges, DCP \cite{he2010single}, AOD-Net \cite{li2017aod}, GridDehazeNet \cite{liu2019griddehazenet}, DehazeFormer \cite{song2023vision} and our proposed RSHazeNet on Haze1K dataset \cite{huang2020single} and RS-Haze dataset \cite{song2023vision}. The subsequent row of each generated image is the corresponding error map.}
    \label{fig:haze1k_rshaze}
    \vspace{-2mm}
\end{figure}

Table \ref{tab:statehaze1k} presents the evaluation results. Our RSHazeNet achieves the best performance in PSNR, SSIM, and MSE on both StateHaze1K and RS-Haze datasets. Specifically, our method outperforms DehazeFormer by an average of 1.61 dB on StateHaze1K dataset and 1.04 dB on RS-Haze dataset.
Furthermore, we provide the visual comparisons, including DCP \cite{he2010single}, AOD-Net \cite{li2017aod}, GridDehazeNet \cite{liu2019griddehazenet}, DehazeFormer \cite{song2023vision} and our RSHazeNet in Figure \ref{fig:haze1k_rshaze}. It can be observed that our proposed method reconstructs haze-free images with the lowest error compared to the other methods.

\subsection{Ablation Studies}
The baseline architecture employ FNB \cite{chen2023run} as the basic learning block and utilizes a $1\times 1$ convolution for intra-level feature fusion. We initially make modifications to the pipeline by adopting an encoder-minimal and decoder-minimal framework (EDF), which results in a slight improvement in performance while simultaneously reducing the computational burden.
As illustrated in Table \ref{tab:component}, our hierarchical framework surpasses the performance of the conventional u-shaped architecture by an improvement of 0.07 dB, while reducing the parameter count by 0.015\% and the FLOPs by 11.83\%.
Subsequently, we undertake an exhaustive analysis of the contributions made by each module, encompassing IFTM, CMIM, MPEB, and soft residual connection (SRC) \cite{song2023vision}. 
Each proposed component shows a positive effect on the overall performance.

\begin{table}[H]\footnotesize
     \centering
     \caption{Ablation study of the individual components. Each proposed component plays a valuable role and shows a positive effect on the overall performance for remote sensing image dehazing.}
     \label{tab:component}
     \renewcommand\arraystretch{1}
     % \resizebox{\linewidth}{!}{
     \setlength{\tabcolsep}{2.6mm}{
     \begin{tabular}{lccccc}
     \toprule
     Model & PSNR & SSIM & MSE & \#Param & FLOPs \\
    \midrule
    \cline{5-6}
    Baseline & 21.95 & 0.8347 & 0.0094 & \multicolumn{1}{|c}{\blue{1.105 M}} & \multicolumn{1}{c|}{\blue{38.89 G}} \\
    + EDF & 22.02 & 0.8494 & 0.0093 & \multicolumn{1}{|c}{\red{1.088 M}} & \multicolumn{1}{c|}{\red{34.29 G}} \\
    \cline{5-6}
    + IFTM & 24.53 & 0.9186 & 0.0072 & 1.113 M & 34.86 G \\
    + CMIM & 25.24 & 0.9258 & 0.0033 & 1.213 M & 40.42 G \\
    + MPEB & \blue{26.12} & \blue{0.9319} & \blue{0.0026} & 1.190 M & 40.07 G \\
    + SRC & \red{26.26} & \red{0.9325} & \red{0.0024} & 1.190 M & 40.14 G \\
    \bottomrule
    \end{tabular}
     }
 \end{table}

\section{Conclusion}

In this work, our proposed RSHazeNet represents a significant advancement in remote sensing image dehazing.
By adopting an encoder-minimal and decoder-minimal architecture and introducing novel modules such as ITFM, CMIM, and MPEB, we have achieved a compelling trade-off between image quality and computational efficiency.
Extensive experimental evaluations on multiple haze-degraded datasets demonstrate that RSHazeNet outperforms the state-of-the-art methods by a substantial margin.

\bibliographystyle{IEEEbib}

\end{document}